\documentclass{article}

\usepackage[final]{neurips_2025}

\usepackage[utf8]{inputenc} 
\usepackage[T1]{fontenc}    
\usepackage{hyperref}       
\usepackage{url}            
\usepackage{booktabs}       
\usepackage{amsfonts}       
\usepackage{graphicx}       
\usepackage{nicefrac}       
\usepackage{microtype}      
\usepackage{xcolor}         
\usepackage{amsmath}
\usepackage{listings}
\title{Limits of Emergent Reasoning of Large Language Models in Agentic Frameworks for Deterministic Games}

\author{%
  Chris Su \\
  School of Computer Science \\
  Carnegie Mellon University \\
  Pittsburgh, PA \\
  \texttt{chrissu@andrew.cmu.edu} \\
  \And
  Harrison Li \\
  College of Computing, Data Science,\\and Society \\
  University of California, Berkeley \\
  Berkeley, CA \\
  \texttt{liharrison@berkeley.edu} \\
  \And
  Matheus Marques \\
  Computer Science \\
  The College of New Jersey \\
  Ewing, NJ \\
  \texttt{marquem1@tcnj.edu} \\
  \And
  George Flint\thanks{Senior author.} \\
  Cognitive Science \\
  University of California, Berkeley \\
  Berkeley, CA \\
  \texttt{georgeflint@berkeley.edu} \\
  \And
  Kevin Zhu\footnotemark[1] \\
  Algoverse AI Research \\
  \texttt{kevin@algoverseacademy.com} \\
  \And
  Sunishchal Dev\footnotemark[1] \\
  Algoverse AI Research \\
  \texttt{dev@algoverseairesearch.org} \\
}

\begin{document}

\maketitle

\begin{abstract}
Recent work reports that Large Reasoning Models (LRMs) undergo a collapse in performance on solving puzzles beyond certain perplexity thresholds. In subsequent discourse, questions have arisen as to whether the nature of the task muddles an evaluation of true reasoning. One potential confound is the requirement that the model keep track of the state space on its own. We provide a large language model (LLM) with an environment interface for Tower of Hanoi problems, allowing it to make a move with a tool call, provide written justification, observe the resulting state space, and reprompt itself for the next move. We observe that access to an environment interface does not delay or eradicate performance collapse. Furthermore, LLM-parameterized policy analysis reveals increasing divergence from both optimal policies and uniformly random policies, suggesting that the model exhibits mode-like collapse at each level of complexity, and that performance is dependent upon whether the mode reflects the correct solution for the problem. We suggest that a similar phenomena might take place in LRMs.\footnote{\textbf{Code: } \url{https://anonymous.4open.science/r/cdz50eodtx/README.md}}

\end{abstract}

\section{Introduction}
Recent advances in large language models (LLMs) have enabled them to solve difficult problems by generating ``thoughts" involved in the solution \citep{wei2022cot, chen2022pot,yao2023tot,besta2024got}. However, these models, often called large reasoning models (LRMs), can struggle with reasoning when faced with problems that require multistep logic \citep{wei2022cot}. This limitation is especially pronounced in domains such as game-like scenarios, e.g. Tower of Hanoi.

There is ongoing discussion that reasoning capability measured by existing benchmarks may be inflated \citep{chollet2019arc,wei2022cot,chen2022pot}, where reasoning ability is often conflated with memorization of training data. Furthermore, existing evaluation approaches of puzzle-solving scenarios require models to maintain internal representations of state spaces, introducing potential confounds that obscure whether failures stem from reasoning deficits or architectural limitations in state tracking. To test this, we provide an environmental framework that externalizes state management. This externalization enables us to conduct policy analysis by treating the model as an agent whose decision patterns can be analyzed. If models possess genuine reasoning capabilities, independent of state management limitations, we should observe policies that approach optimal behavior, differ meaningfully from random exploration, and exhibit exploratory patterns when faced with novel states. Conversely, if models exhibit deterministic behavior, this would suggest that apparent reasoning is actually fixed, learned patterns rather than dynamic problem-solving capabilities. We hypothesize that allowing models to interact dynamically with problem states rather than maintaining internal representations will force the model to use emergent heuristics and reasoning capabilities outside of the model's priors created during the training process.

To this end, we analyze criticism of LLM and LRM reasoning capabilities \citep{shojaee2024, lawsen2025} through the lens of an agentic environment framework, where the foundational model must plan and execute a series of optimal steps that will allow it to reach the goal state. We build on an increasing complexity approach introduced by \citet{shojaee2024}, extending an interactive environment where models can make sequences of singular tool calls. This setup allows us to track the reasoning states of the models and observe their decision making in a structured manner. Our makes two key contributions to understanding emergent reasoning in large language models. First, we establish that providing dynamic environment interfaces does not mitigate performance collapse in reasoning tasks. Second, we demonstrate that models increasingly diverge from optimal policies and uniformly random policies, suggesting that the models are incapable of learning from past mistakes. In particular, we find that the models fall into looping behavior, characterized by sequences which transitions return to previously visited states. For deterministic games with single goal states such as Tower of Hanoi, this is considered un-optimal or problematic behavior, as returning to a previously visited state implies that the agent has made no progress.

\section{Related Works}

\citet{shojaee2024} proposed a puzzle-based framework featuring controllable problem complexity to evaluate Large Language Model (LLM) and Large Reasoning Model (LRM) reasoning. The paper demonstrated that LLMs and their reasoning counterparts both systematically collapse at high complexity. Critics of this work argue that observed failures stem from experimental limitations, such as token limits or unsolvable problems\citep{lawsen2025}. Reasoning performance is noted to be highly task-dependent and nonlinear, and the debate on reasoning in LRMs is nuanced\citep{varela2025rethinking}. 

While there is a diverse landscape of reasoning benchmarks, this proliferation shows the difficulty of defining reasoning itself. The concept of emergent reasoning, where qualitatively new abilities appear at certain model scales\citep{wei2022emergent,berti2025emergent}, has gained particular attention. Models trained on structured data like code have shown improved generalization across varied tasks\citep{aryabumi2024code}. Yet this apparent emergence raises questions: Are models developing actual reasoning capabilities? Or merely becoming more adept at recognizing and reproducing logical patterns from training data?

Our approach builds directly on the  complexity-based evaluation of \citet{shojaee2024} while also addressing experimental limitations and providing a framework that separates reasoning assessment from potential confounds and allows for analysis of LLM policy.

\section{Methodology}
We evaluate reasoning as search in the context of the Tower of Hanoi puzzle. This environment requires systematic exploration and move pruning, as well as a controllable search space where search space size scales exponentially with increasing number of disks \citep{shojaee2024, lawsen2025}. 

\subsection{Tower of Hanoi}

The Tower of Hanoi puzzle is a recursive puzzle consisting of three pegs and $n$ disks of distinct sizes initially stack in order of size on a single peg (largest on the bottom). The objective is to transfer the entire stack from the first peg to the third peg. The rules are (1) that only the top disk of a peg may be moved, and (2) that a larger disk may never be placed on top of a smaller one. The difficulty of this puzzle grows with the number of disks, where the optimal number of moves required to solve an n-disk instance is $2^n - 1$ \citep{frame1941tower}.

\subsection{Experimental Setup}

\subsubsection{Baseline}
We first establish a baseline in which models are required to generate a complete Tower of Hanoi solution in a single pass with no interaction with the environment. Our experiments are conducted on large language models and their LRM counterparts. Similar to the experimental setup of \citet{shojaee2024}, we specifically use Claude 3.7 Sonnet (and its reasoning counterpart), DeepSeek V3.1, and DeepSeek R1 because these allow access to reasoning traces\citep{anthropic2025claude, deepseek2025v3}. 

We restrict evaluation to up to n = 8 disks, which corresponds to the regime in which IOT reported almost complete collapse. Subsequent critiques of that work also noted that several of the environments they used contained unsolvable or ill-defined instances \citep{lawsen2025}, making it difficult to separate true model failures from task design artifacts. \citet{lawsen2025} argue that the observed maximum $n$ before collapse was largely due to token budget limitations, rather than a fundamental limit of model reasoning capacity. To avoid these confounds, we focus exclusively on Tower of Hanoi, which is both widely studied and guaranteed to be solvable for all problem sizes. This ensures that any observed model behavior directly reflects reasoning performance, rather than inconsistencies in the environment specification. 

Following \citet{lawsen2025}, the number of tokens required to represent a Tower of Hanoi trajectory increases approximately quadratically, as the evaluation format requires the full move sequence at each step. Assuming about 5 tokens per move, this can be expressed as:
\[
T(N) \approx 5 \cdot (2^N - 1)^2 + C
\]
With a budget of 64,000 tokens, this yields maximum solvable sizes of:
\[
N_{\text{max}} \approx \left\lfloor \log_2 \left( \frac{L_{\text{max}}}{5} \right) \right\rfloor
\]

\[
 \approx 7\text{--}8 \quad \text{(Claude 3.7, DeepSeek V3.1)}
\]
roughly 7–8 disks for Claude 3.7 and DeepSeek V3.1. 

The maximum budget of 64,000 tokens was not enforced. Instead, the token budget was set to 30,000 tokens, which was sufficient based on the output lengths reported in \citet{shojaee2024}. Under this setting, models never display truncation, which suggests that practical token usage fell below the quadratic growth curve identified in \citet{shojaee2024}, and token constraints were not the dominant limiting factor. 

\subsubsection{Environment-Based Agentic Framework}
\begin{figure}[h]
    \centering
    \includegraphics[width=1\linewidth]{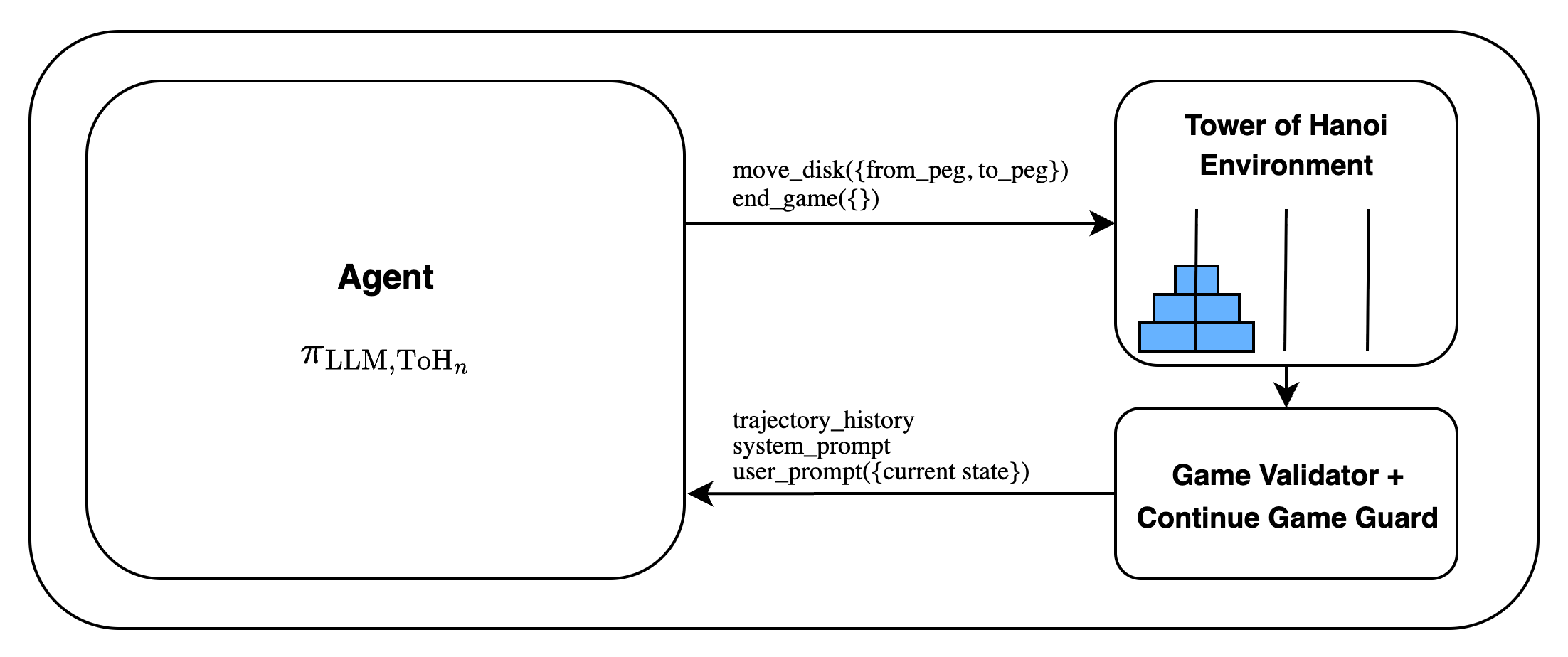}
    \caption{Agentic framework is a closed loop interaction between the agent, environment, and game validator.}
    \label{fig:placeholder}
\end{figure}
We implement an interactive evaluation framework in which the model engages directly with the Tower of Hanoi environment through tool calls. Rather than producing an entire solution trajectory in a single generation, the model must act step by step and incrementally explore the state space.

We connect Claude 3.7 and DeepSeek V3.1 to the environment, which exposes  two API endpoints: (1) \texttt{move\_disk(from\_peg, to\_peg)}, which moves the top disk between pegs (invalid moves are blocked by the environment); and (2) \texttt{end\_game()}, which allows the model to terminate the run.

At each step, the model is given the system prompt, user prompt, and full history of prior moves. The environment responds with structured feedback after every tool call, such as the new state.

The baseline requires the model to produce an entire solution trajectory in a single generation. This format may advantage models by allowing them to rely on distributional memorization of optimal trajectories present in training data, rather than maintaining reasoning over long horizons. The agentic framework requires the model to reason incrementally where each move is conditioned on the evolving puzzle state and prior actions, preventing direct retrieval of complete solutions.
\subsection{LLM-Parameterized Policy Analysis}
We employed Q-Value policy analysis to observe how the LLM prunes and takes actions when in some state, $s$. Namely, we used a useful theorem available in Appendix \ref{q-value-convergence} which states that Q-values for state-action pairs converge to a function which implicitly maps distance from goal state when positive rewards are given to finding the goal state in deterministic, single, absorbing goal state, search games, such as Tower of Hanoi.

This formalization allows an easier way to implement a way to find state-action pair equivalences that classical search algorithms have difficulty formalizing.

We sampled trajectories (a sequence of transitions) from the LLM's playout of varying Tower of Hanoi instances and constructed a dataset $\mathcal{D} = \{(s_i, a_i, s_i')\}_{i=1}^N$. Given some state, $s$, we can parameterize a search policy, $\pi$, against the LLM's moves using the proposed agentic framework as
$$\pi_{\text{LLM},\text{ToH}_n}(s, a) = \mathbb{P}\big[S=s,A=a\big]$$
Which can be approximated as
$$\hat\pi_{\text{LLM},\text{ToH}_n}(s,a)=\frac{\sum_{i=1}^N\mathbf{1}[s_i=s,a_i=a]}{N}$$
Note that by Bayes Theorem, this policy can be easily transformed to the more standard conditional form $\pi(a | s)$, but we choose to use this joint form for convenience. This policy can be decomposed to "sub"-policies, $\pi_2, \pi_3$, which are probability distributions defined only for states which have 2 and 3 valid actions, respectively:
\[
\hat \pi_{\text{LLM},\text{ToH}_n}(a|s) = 
\begin{cases}
\hat \pi_2(s, a \big| \text{size}(V(s))=2), & \text{if } |V(s)|=2 \\
\hat \pi_3(s, a \big|\text{size}(V(s))=3), & \text{otherwise}
\end{cases}
\]
Note that $\mathcal{S}_n$ is the set of states of Tower of Hanoi with $n$ disks, $V(s) := \{\text{set of valid actions from } s\}$. Also observe that every (valid) ToH state has two valid moves or three valid moves. That is, $\forall n, \forall s \in \mathcal{S}_n, 2 \leq |V(s)| \leq 3$. However, the only nontrivial case occurs when $|V(s)| = 3$ since a ToH state has two valid actions if and only if all disks are on the same peg (see Appendix \ref{toh-bounds}). We used the distribution of $\hat \pi_3$ to compare against the following agents: (1) optimal agent, (2) random agent. For some environment, $E$, we define a random agent as any agent for which its policy comes from the following distribution, for all states in the environment

$$\pi(s) \sim U(V(s)), \text{ } \text{ }\forall s\in \mathcal{S}_E$$

where $U(\cdot)$ is the uniform distribution. For some environment, $E$, we also define an optimal agent as any agent for which its policy is "optimal". Namely $$\pi^*(s) := \text{argmax}_a Q^*(s, a), \forall s \in \mathcal{S}_E$$

We used the Jensen-Shannon divergence as our primary metric of comparison to the optimal and random agent policies. For distributions $P, Q$, Jensen-Shannon Divergence (JSD) is defined as 
$$JSD(P || Q)=\frac{1}{2}KL(P||V) + \frac{1}{2}KL(Q||V)$$

$$V = \frac{1}{2}(P+Q)$$

The Jensen-Shannon divergence has useful properties such as being bounded between [0, 1] and being symmetric ($JSD(P||Q) = JSD(Q||P)$) (see Appendix \ref{JSD-bounds}). $JSD(P||Q)$ can be interpreted as the similarity between $P$ and $Q$, where $JSD(P||Q) = 0$ means that $P$ and $Q$ are identical and $JSD(P||Q)  = 1$ means that $P$ and $Q$ have different supports. To ensure interpretable JSD comparison between differing levels of complexity, we conditioned the optimal policy on whether the state was also visited by the model during its exploration.

\section{Results}
\subsection{Baseline}
\begin{figure}[h]
    \centering
    \includegraphics[width=1\linewidth]{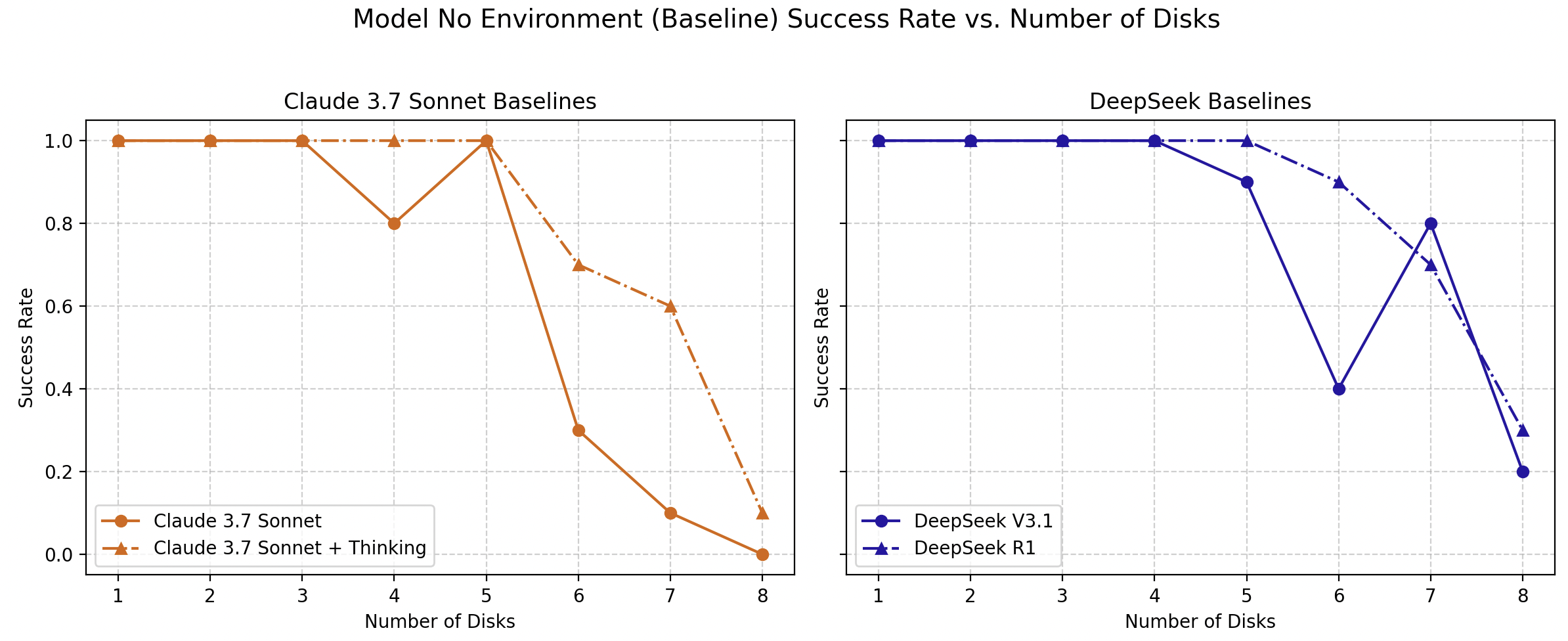}
    \caption{Comparison of success rates of LLM and LRM one-shot generation: (Left) Claude 3.7 Sonnet with and without "thinking" mechanism, (Right) DeepSeek V3.1 vs R1. Line charts display success rate as a function of puzzle complexity.}
    \label{fig:baselines-combined}
\end{figure}
Figure~\ref{fig:baselines-combined} shows that as complexity increases, all models, thinking or non-thinking, display a similar pattern: success rates collapse (near-zero accuracy) when crossing a certain threshold. LRMs consistently outperform non-thinking models, but still succumb to collapse at high complexity. This finding replicates prior reports of performance collapse in LRMs \citep{shojaee2024}. 

\subsection{Agentic Framework}
\begin{figure}[h]
    \centering
    \includegraphics[width=1\linewidth]{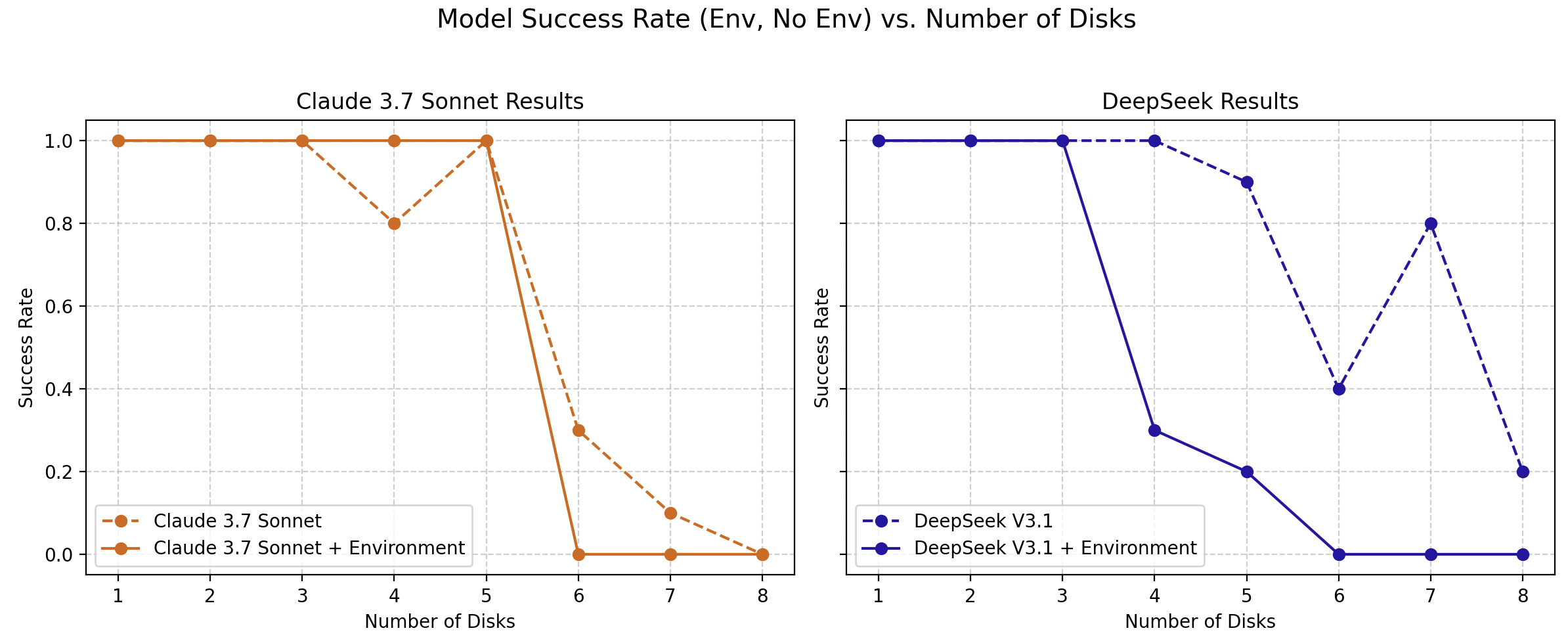}
    \caption{Success rate of models in an agentic framework (Claude 3.7 Sonnet + environment, DeepSeek V3.1 + environment) in comparison to the baseline (Claude 3.7 Sonnet, DeepSeek V3.1) at increasing complexity levels.}
    \label{fig:baseline-agent}
\end{figure}

\begin{figure}[h]
    \centering
    \includegraphics[width=1\linewidth]{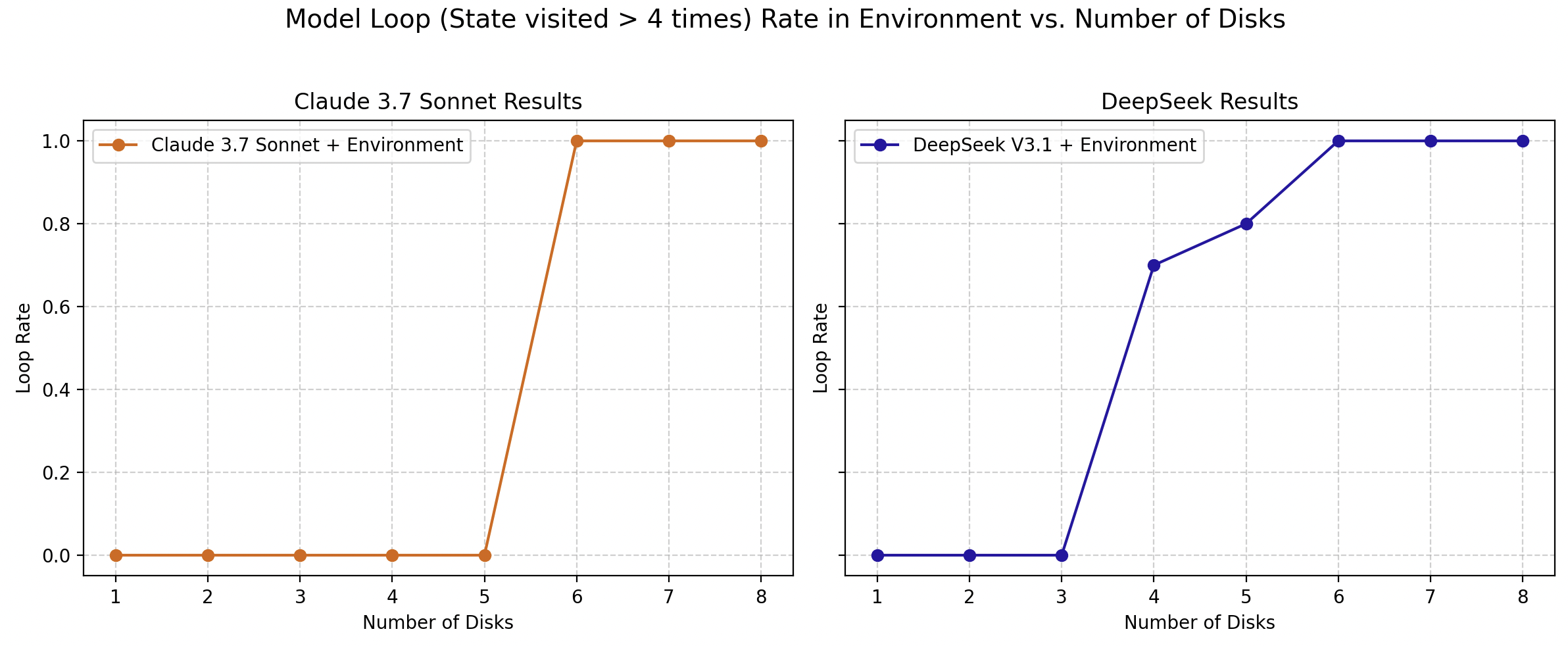}
    \caption{Loop rate of the models in an agentic framework (Claude 3.7 Sonnet + environment, DeepSeek V3.1 + environment) at increasing complexity levels.}
    \label{fig:loop-rate}
\end{figure}

\begin{figure}
    \centering
    \includegraphics[width=1\linewidth]{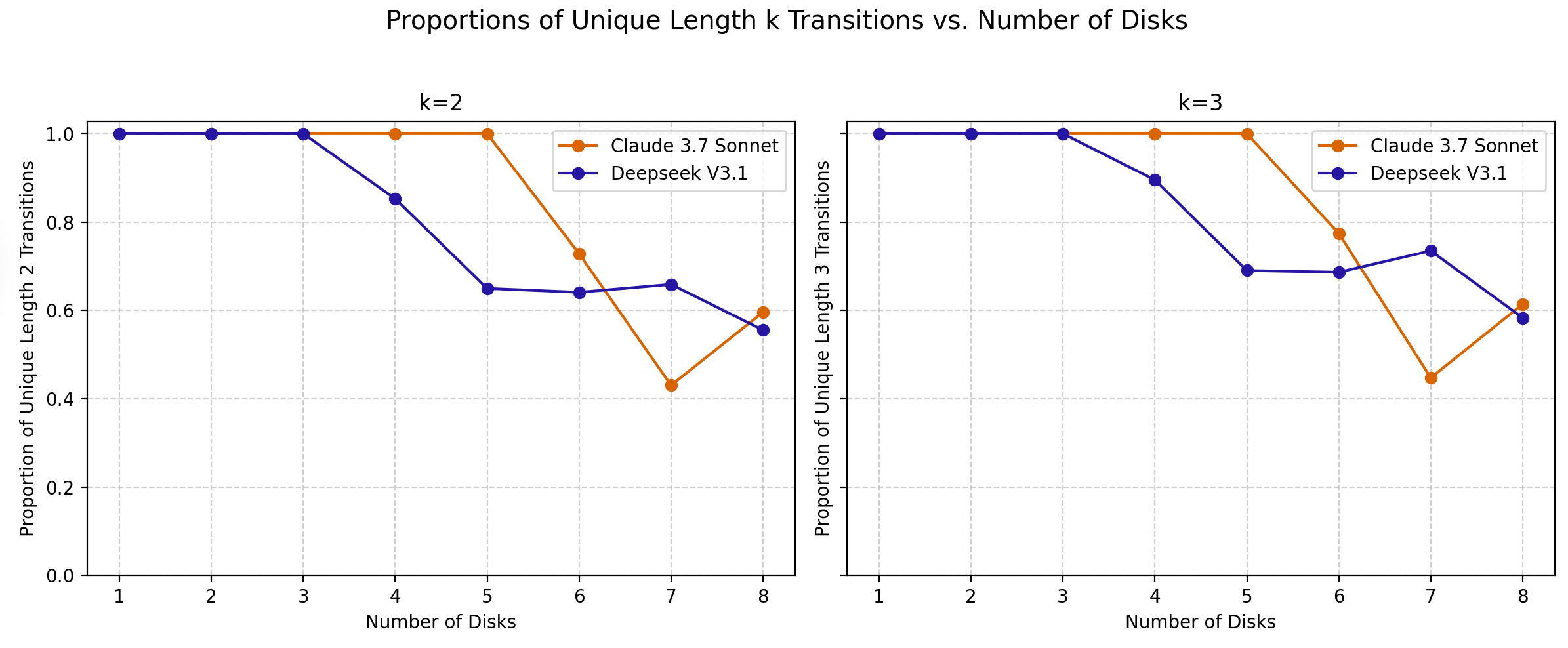}
    \caption{Proportion of unique length $k$ transitions taken from state $s$, given that $s$ was visited by the model at least twice. Lower values mean that the model takes less unique length $k$ trajectories. These graphs are similar since every $k=3$ subsequence from $s$ also contains the $k=2$ subsequence from $s$.}
    \label{fig:subsequence}
\end{figure}

The results in Figure~\ref{fig:baseline-agent} show that the introduction of an environment interface does not prevent performance collapse. Instead, degradation occurs at a lower complexity than in the baseline. Under an agentic framework, collapse is associated with failing to escape deterministic looping behavior (Figure~\ref{fig:loop-rate}). Subsequence analysis in Figure~\ref{fig:subsequence} shows about 40\% of the time the model deterministically reuses previously observed continuations at $n=8$. Unique length $k$ transitions is observed to decrease at moderate-high N.

\subsection {LLM-Parameterized Policy Analysis}
\begin{figure}[h]
    \centering
    \includegraphics[width=1\linewidth]{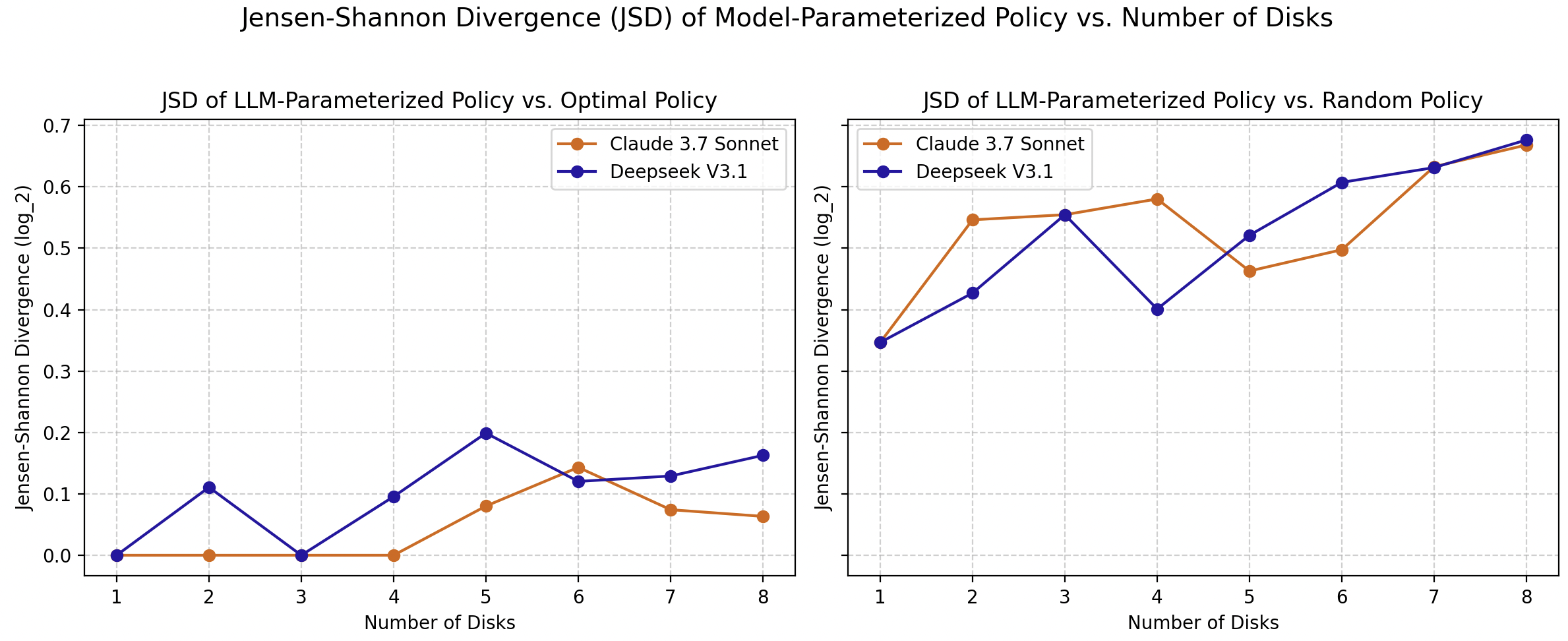}
    \caption{Jensen Shannon Divergences of LLM-Parameterized policies against Optimal policies and Random policies.}
    \label{fig:JSD}
\end{figure}
With increasing complexity, we see two behaviors shown in Figure~\ref{fig:JSD}. The first is that the model-parameterized policy diverges from the optimal policy. The second is that the model-parameterized policy also diverges from the uniformly random policy. We note that collapse is associated with these two behaviors. 

\section{Discussion}
Although the model is given access to its full move history and current state, stepwise execution exposes it to sparse intermediate states. In these cases, models exhibit deterministic behavior collapse, often becoming locked into suboptimal action loops. The agentic framework performs worse than baseline models, which could suggest stepwise interaction actually exacerbates rather than mitigates underlying reasoning limitations. The agentic setting reveals characteristics of brittle reasoning that the baseline obscures. 

This is most clear in Jensen-Shannon Divergence (JSD) analysis of the respective policies. The JSD of the optimal policy against the observed LLM parameterized policy noticeably increases with problem complexity, meaning models diverge progressively from optimal behavior with increasing $n$. Furthermore, JSD between LLM policy and random policy also increases with complexity, suggesting some of the probability weights, which are uniform for the random agent, are being reassigned unequally, suggesting that the model has priors about what actions to take or prune in certain states. The simultaneous divergence from both optimal and random policies implies that models are neither reasoning optimally nor exploring effectively. Instead, they are executing deterministic patterns that become increasingly maladaptive. 

Subsequence analysis provides possible evidence of this deterministic pattern execution. When models return to previously visited states, they consistently execute identical suboptimal action sequences. This repetition occurs despite models having access to their complete interaction history. Hence, this repetition for lower N values can imply being deterministically correct, while at high N, its tendency to decrease shows signs of deterministic patterns through revisiting the same unique states. This inability to vary its behavior upon encountering familiar states, even after experiencing negative consequences from identical previous trajectories, could imply that apparent reasoning is actually execution of fixed computational patterns.





The model's inability to generalize and learn from its history, incorporate long-term planning, and correct its own priors seems to affect the model's ability to perform in dynamic environments which have interactions that are outside of the model's training distribution. 

\paragraph{Limitations.} Our investigation only examines two LLMs--Claude 3.7 Sonnet and Deepseek V3.1--and their LRM counterparts--Claude 3.7 Sonnet Thinking and Deepseek R1. Likewise, we only examine one game task--Tower of Hanoi. It is unclear how our results might generalize to other classes of tasks or models. We did not conduct repeated trials per model, so results reflect single-run outcomes and may be sensitive to sampling variance, though \citet{shojaee2024} report limited variance in their trials. We additionally do not report any ablations over temperature, self-consistency, majority vote, or beam search, which could shift results.

\section{Conclusion}
We examine the reasoning performance of Large Language Models in both single-pass and agentic environments. Across both, performance collapse emerged as a consistent pattern, and demonstrated that environment access does not delay or prevent performance collapse. Instead, models exhibit deterministic adherence to a single behavioral trajectory, with success hinging on whether the trajectory matched the solution. Taken together, our findings reinforce that apparent reasoning ability is largely a byproduct of high-probability mode following, rather than genuine reasoning. More broadly, this work adds to growing evidence that scaling alone is insufficient in creating general-purpose emergent reasoning capabilities in large language models.



\bibliographystyle{unsrtnat}

\section*{Appendix}
\appendix

\section{Theorem: Q-Value Convergence in Single Goal, Strongly Connected Deterministic Games} \label{q-value-convergence}

Let $G$ be a single goal state, strongly connected, deterministic game. Let $\mathcal{S}$ be the state space for $G$. Let $\mathcal{G}$ be the singleton set that contains the goal state of $G$, such that $\mathcal{G} = \{s \in \mathcal{S} : s \text{ is an absorbing goal state\}}$, with $|\mathcal{G}| = 1$ and the game ends when the (absorbing-)goal state is achieved. Let $\mathcal{A}$ be the action space. Let $T: \mathcal{S} \times \mathcal{A} \rightarrow \mathcal{S}$ be the transition function. Since the game is deterministic, we have that $P(s' |s, a) = 1$. Let $R: \mathcal{S} \times {A} \times \mathcal{S} \rightarrow \mathbb{R}$ be a function that maps state, action, and next state to some reward, $r$. We impose the following condition on $R$:

\[
R(s,a,s') =
\begin{cases}
1, & \text{if } s' \in \mathcal{G} \\
0, & \text{otherwise}
\end{cases}
\]

Suppose that the agent observes a transition $(s, a, r, s')$, with $r = R(s, a, s')$.

Let $\alpha \in \mathbb{R}$, $\gamma \in \mathbb{R}$, with $0< \gamma < 1$. Let $H^{*}(s) : \mathcal{S}\rightarrow\mathbb{N}$, such that 
\[
H^{*}(s) =
\begin{cases}
0, & \text{if } s\in \mathcal{G} \\
\min_{p \in P(s, g), g\in\mathcal{G}} |p|, & \text{otherwise}
\end{cases}
\]
where $P(u, v):= \{p | p \text{ is a path that starts at } u \text{ and ends at } v\}$. 
Finally, suppose that the agent acts randomly with probability $0 < \epsilon < 1$ and with probability $1 - \epsilon$, follows the current learned policy, $\pi(s) = \text{argmax}_{a \in V(s)}Q(s,a)$, where $V:\mathcal{S} \rightarrow \mathcal{P}(\mathcal{A})$ is a function mapping states, $s$ to a set of valid actions $a$ from $s$. We show that after convergence of $Q \hookrightarrow Q^*$, which follows from an appropriately chosen $\alpha$, the Bellman Optimality Equation
\[
Q^*(s,a) = \mathbb{E} \left[ r + \gamma \max_{a'} Q^*(s',a') \,\middle|\, s,a \right]
\]
is satisfied by a function with the following characteristics:
\begin{align}
\hat Q(s, a_1) = \hat Q(s, a_2) \iff H^{*}(T(s, a_1)) = H^{*}(T(s, a_2)) \\ H^{*}(T(s, a_1)) > H^{*}(T(s, a_2)) \Rightarrow \hat Q(s,a_1) < \hat Q(s, a_2)
\end{align}

\textbf{Proof.} Intuitively, the proof arrives from the idea that with sufficient exploration (large enough $N$), and with some discount factor, $\gamma$ less than 1, the reward of being in a single goal state will propagate throughout the state space graph. Each additional transition, $t$, away from the goal state adds multiplies another factor of $\gamma$ discount to the value of taking $t$. 

Observe that transitions are deterministic. Hence, the Bellman Optimality Equation can be simplified to the following

\begin{align}
Q^*(s,a)
&= \sum_{s' \in V(s)}P(s' |s, a)\left[ R(s, a, s') + \gamma \max_{a'} Q^*(s',a') \,\middle|\, s,a \right] \\
&= R\bigl(s,a,T(s,a)\bigr) + \gamma\max_{a'} Q^*\bigl(T(s,a),a'\bigr) \\
&= \mathbf{1}_{\{T(s,a)\in\mathcal G\}} + \gamma\max_{a'} Q^*\bigl(T(s,a),a'\bigr)
\end{align}

We claim that $Q^*$ has closed form: $Q^*(s, a) = \hat Q(s, a) = \gamma^{H^{*}(T(s, a))}$. These functions must thus agree on all valid tuples $(s, a) :  s \in \mathcal{S},  a \in V(s)$. We have two cases to consider. Namely, the case where $T(s, a) \in \mathcal{G}$ and the case where $T(s, a) \notin \mathcal{G}$. In the first case, the Bellman Optimality Equation satisfies
\begin{align}
Q^*(s,a) 
&= 1 + \gamma\max_{a'} Q^*\bigl(T(s,a),a'\bigr) \\
&= 1 + \gamma * 0 \\ 
&= 1
\end{align}

Observe that $\hat Q(s, a) = \gamma^{H^{*}(T(s, a))} = \gamma^{0} = 1$, as desired. Note that $H^*(T(s,a)) = 0$, by definition of $H^*$. For the second case, the Bellman Optimality Equation satisfies
\begin{align}
Q^*(s,a) 
&= 0 + \gamma\max_{a'} Q^*\bigl(T(s,a),a'\bigr) \\
&=  \gamma \max_{a'} Q^*\bigl(T(s,a),a'\bigr)
\end{align}

Assume for the sake of contradiction that $\hat Q$ did not satisfy the Bellman Optimality Equation. That is

\begin{align}
\hat Q(s, a) \neq \gamma \max_{a'} \hat Q\bigl(T(s,a),a'\bigr)
\end{align}

Plugging in our definition for $\hat Q$ into (11), we get the following inequality

\begin{align}
\gamma^{H^*(T(s,a))} 
&\neq \gamma \max_{a'} \gamma^{H^*(T(s, a'))}
\end{align}
Recall that the state space graph is strongly connected, which implies that there exists some path from $s$ to $g \in \mathcal{G}$. The key insight is to consider the shortest path $p$ from $s' = T(s,a)$ to the goal state, $g$. Without loss of generality, suppose that ties in length of $p$ candidates are broken in some arbitrary (but initially chosen) direction. Let $s''$ be the next state from $s'$ in this path $p$. By definition of $H^*$, we must have

$$H^*(s')=H^*(s'') + 1$$

since both states, $s'$ and $s''$ lie on the shortest path to $g$. Under the same tie breaking scheme as before, consider the term
\begin{align}
\max_{a'} \gamma^{H^*(T(s, a'))}
\end{align}

which is maximized only by taking the transition, $T(s, a') = s''$. Hence it follows that $\text{argmax}_{a'}H^*(T(s,a')) = s''$ implying 

\begin{align*}
\max_{a'} \gamma^{H^*(T(s, a'))} 
&= \gamma^{H^*(s'')} \\
&= \gamma^{H^*(T(s, a'))} \\
&= \gamma^{H^*(s') - 1}
\end{align*}
Substituting back into the inequality (12) we get
\begin{align}
\gamma^{H^*(T(s,a))} 
&\neq \gamma \max_{a'} \gamma^{H^*(T(s, a'))} \\ 
&\neq \gamma * \gamma^{H^*(s') - 1} \\ 
&\neq \gamma^{H^*(s')} \\
&\neq \gamma^{H^*(T(s,a))}
\end{align}
a contradiction.

Observe that this definition for $\hat Q$ satisfied properties $(1)$ and $(2)$. Property $(1)$ follows immediately as $\gamma^{H^{*}(T(s, a_1))} = \gamma^{H^{*}(T(s, a_2))} \iff H^{*}(T(s, a_1)) = H^{*}(T(s, a_2))$. Property (2) also follows immediately as $0 < \gamma < 1$, so $\gamma ^ {k} > 0, k \in \mathbb{R}$. Hence $$H^{*}(T(s, a_1)) < H^*(T(s, a_2)) \Rightarrow \gamma^{H^{*}(T(s, a_1))} > \gamma^{H^{*}(T(s, a_2))}$$

\section{Theorem: Valid Action Set is Bounded by [2, 3] for Tower of Hanoi} \label{toh-bounds}
Let $V(s) = \{a : a \text{ is a valid action from s}\}$. Let $\mathcal{S}_n$ be the state space of $n$ disk ToH. Then, $\forall n, s\in \mathcal{S}_n, 2\leq|(V(s)|\leq3$. There are three cases to consider. Namely, when the state space has one, two, and three pegs occupied.

\textbf{Case 1.}
Let $s$ be a state such that it has only one peg occupied. Without loss of generality, say that peg $1$ is occupied. Since there are no disks on any other peg, it must be the case that the disk on peg $1$ can be moved to peg $2$ or peg $3$. Hence $|V(s)| = 2$.

\textbf{Case 2.}
Let $s$ be a state such that it has two pegs occupied. Without loss of generality, say that peg $1$ and peg $2$ are occupied. Further, without loss of generality, suppose that $d_1 < d_2$, where $d_i$ is the disk on peg $i$. Observe that both $d_1$ and $d_2$ can be moved to (the empty) peg $3$. Further, since $d_1 < d_2$, we may move $d_1$ to peg $2$. Hence, $|V(s)| = 3$.

\textbf{Case 3.}
Let $s$ be a state such that it has three pegs occupied. Without loss of generality, say that $d_1 < d_2 <d_3$. This ordering exists since all disks are unique. Observe that $d_1$ may be moved to peg $2$ and peg $3$, and $d_2$ may be moved to peg $3$, since the ordering is satisfied. Hence, $|V(s)| = 3$.

\textbf{Conclusion}
This concludes the proof. However, we note that the first case says something stronger about the valid action set of some arbitrary state. Namely $$|V(s)| = 2 \iff s \text{ is the goal state, or identical (in ToH progress) to the initial state}, s_o$$

This fact allows us to exclude this case from LLM-parameterized policy analysis.

\section{Theorem: Jensen-Shannon Divergence is Bounded by [0, 1] and Symmetric} \label{JSD-bounds}
Let $P$, $Q$ be arbitrary (discrete) distributions on elements $x \in \chi$. Use the convention $0*\log(0) = 0$. Define the mixture, $M(P,Q) := \frac{1}{2}(P+Q)$. We show that

$$JSD(P||Q) = \frac{1}{2}KL(P||M(P,Q))+\frac{1}{2}KL(Q||M(P,Q))$$

has two properties. (1) $0 \leq JSD(P||Q) \leq 1$ and (2) $JSD(P||Q) = JSD(Q||P)$.

\textbf{Property 1.} Recall $KL(\cdot || \cdot) \geq 0$. Hence, the left side of the inequality holds. That is, $JSD(P||Q) \geq 0$. To show that $JSD(P||Q) \leq 1$, we expand the definitions of $KL$ divergence to get

$$KL(P||M) = \sum_{x\in \chi}P(x)\log\frac{P(X)}{M(x)}$$

Observe that the following holds:

$$M(x)=\frac{P(x) + Q(x)}{2} \geq \frac{P(x)}{2}, \forall x\in \chi$$

Hence, $\frac{P(x)}{M(x)}\leq 2$ and $\log_2\frac{P(x)}{M(x)}\leq 1$. Plugging this back into our definition for $KL$ divergence, we get

$$KL(P||M) \leq\sum_{x\in\chi}P(x)\cdot 1 = 1$$
The same applied for $KL(Q||M(P,Q)$, and thus we may upper bound $JSD$ as

$$JSD(P||Q) \leq \frac{1}{2}\cdot1 + \frac{1}{2} \cdot 1 = 1$$

as desired.

\textbf{Property 2.} The symmetric nature of $JSD(P||Q)$ follows immediately from its definition, as $M(P,Q) = M(Q,P)$. 

\section{Prompt Design}
\begin{lstlisting}[caption={System Prompt — Tower of Hanoi, Agentic}]
You are a helpful assistant. Solve this puzzle for me step by step using the provided tools.
        
        There are three pegs and n disks of different sizes stacked on the first peg. The disks are numbered from 1
        (smallest) to n (largest). Disk moves in this puzzle should follow:
            
            1. Only one disk can be moved at a time.
            2. Each move consists of taking the upper disk from one stack and placing it on top of another stack.
            3. A larger disk may not be placed on top of a smaller disk.
            The goal is to move the entire stack to the third peg.
            
        Solution Strategy:
            1. Make moves using move_disk(from_peg, to_peg) until you reach the goal state
            2. When you think you've reached the goal state, end the game using end_game().
\end{lstlisting}
\begin{lstlisting}[caption={User Prompt — Tower of Hanoi, Agentic}]
I have a puzzle with {num_disks} disks (numbered 1, 2, 3, ... from smallest to largest) of different sizes with
    - There are 3 pegs (numbered 0, 1, 2).
    - You will start in the Initial state: {initial_state}
    - Want to end in the Goal state: {goal_state}

    Example (3-disk puzzle):
    Initial State: [[3, 2, 1], [], []]
    Goal State: [[], [], [3, 2, 1]]
    Example Solution:
    1. move_disk(0, 2)  
    2. move_disk(0, 1)   
    3. move_disk(2, 1)  
    4. move_disk(0, 2)  
    5. move_disk(1, 0)  
    6. move_disk(1, 2)  
    7. move_disk(0, 2)  

    (This means: Move disk 1 from peg 0 to peg 2, then move disk 2 from peg 0 to peg 1, and so on.)

    Solve the puzzle using the available tools. Move step by step until you reach the goal.


\end{lstlisting}

\newpage
\section*{NeurIPS Paper Checklist}

\begin{enumerate}

\item {\bf Claims}
    \item[] Question: Do the main claims made in the abstract and introduction accurately reflect the paper's contributions and scope?
    \item[] Answer: \answerYes{}
    \item[] Justification: Yes. We outline what we are to show and faithfully detail our process of showing it in the body of the MS.

\item {\bf Limitations}
    \item[] Question: Does the paper discuss the limitations of the work performed by the authors?
    \item[] Answer: \answerYes{}
    \item[] Justification: {This is a subsection of Discussion.}

\item {\bf Theory assumptions and proofs}
    \item[] Question: For each theoretical result, does the paper provide the full set of assumptions and a complete (and correct) proof?
    \item[] Answer: \answerYes{}
    \item[] Justification: {We provide all proofs within the Appendix.}

    \item {\bf Experimental result reproducibility}
    \item[] Question: Does the paper fully disclose all the information needed to reproduce the main experimental results of the paper to the extent that it affects the main claims and/or conclusions of the paper (regardless of whether the code and data are provided or not)?
    \item[] Answer: \answerYes{}
    \item[] Justification:{Yes, within the appendix, and the code is fully provided.}

\item {\bf Open access to data and code}
    \item[] Question: Does the paper provide open access to the data and code, with sufficient instructions to faithfully reproduce the main experimental results, as described in supplemental material?
    \item[] Answer: \answerYes{}
    \item[] Justification: {Yes, the code was linked under the introduction for reproduction.}

\item {\bf Experimental setting/details}
    \item[] Question: Does the paper specify all the training and test details (e.g., data splits, hyperparameters, how they were chosen, type of optimizer, etc.) necessary to understand the results?
    \item[] Answer: \answerNA{}{} 
    \item[] Justification: {We either do not use or do not deviate from default values for these experimental setting values.}

\item {\bf Experiment statistical significance}
    \item[] Question: Does the paper report error bars suitably and correctly defined or other appropriate information about the statistical significance of the experiments?
    \item[] Answer: \answerNo{}
    \item[] Justification: {Our investigation did not include multiple trials, so no error bars were reported.}

\item {\bf Experiments compute resources}
    \item[] Question: For each experiment, does the paper provide sufficient information on the computer resources (type of compute workers, memory, time of execution) needed to reproduce the experiments?
    \item[] Answer: \answerNo{}
    \item[] Justification: {Resources required for running (primarily API usage) are evident int he repository, though we do not mention in the manuscript.}
    
\item {\bf Code of ethics}
    \item[] Question: Does the research conducted in the paper conform, in every respect, with the NeurIPS Code of Ethics \url{https://neurips.cc/public/EthicsGuidelines}?
    \item[] Answer: \answerYes{}
    \item[] Justification: {The paper follows the NeurIPS code of ethics. To our best approximation, there were no potential harms.}

\item {\bf Broader impacts}
    \item[] Question: Does the paper discuss both potential positive societal impacts and negative societal impacts of the work performed?
    \item[] Answer: \answerNA{}
    \item[] Justification: {No parts of our investigation warrant any significant nor direct societal impact.}
    
\item {\bf Safeguards}
    \item[] Question: Does the paper describe safeguards that have been put in place for responsible release of data or models that have a high risk for misuse (e.g., pretrained language models, image generators, or scraped datasets)?
    \item[] Answer: \answerNA{}
    \item[] Justification: {There was no reasonable risk of misuse of any materials involved in our investigation. Standard safeguards for models used are available in their respective model cards.}

\item {\bf Licenses for existing assets}
    \item[] Question: Are the creators or original owners of assets (e.g., code, data, models), used in the paper, properly credited and are the license and terms of use explicitly mentioned and properly respected?
    \item[] Answer: \answerNA{}
    \item[] Justification: {Not applicable because the paper does not use external assets.}

\item {\bf New assets}
    \item[] Question: Are new assets introduced in the paper well documented and is the documentation provided alongside the assets?
    \item[] Answer: \answerNA{}
    \item[] Justification: {Not applicable because the paper does not use external assets.}

\item {\bf Crowdsourcing and research with human subjects}
    \item[] Question: For crowdsourcing experiments and research with human subjects, does the paper include the full text of instructions given to participants and screenshots, if applicable, as well as details about compensation (if any)? 
    \item[] Answer: \answerNA{}
    \item[] Justification: {Our investigation does not have human subjects and does not involve crowd sourcing.}

\item {\bf Institutional review board (IRB) approvals or equivalent for research with human subjects}
    \item[] Question: Does the paper describe potential risks incurred by study participants, whether such risks were disclosed to the subjects, and whether Institutional Review Board (IRB) approvals (or an equivalent approval/review based on the requirements of your country or institution) were obtained?
    \item[] Answer: \answerNA{}
    \item[] Justification: {No IRB approval was required for our investigation.}

\item {\bf Declaration of LLM usage}
    \item[] Question: Does the paper describe the usage of LLMs if it is an important, original, or non-standard component of the core methods in this research? Note that if the LLM is used only for writing, editing, or formatting purposes and does not impact the core methodology, scientific rigorousness, or originality of the research, declaration is not required.
    \item[] Answer: \answerYes{}
    \item[] Justification: {There was usage of LLMs (Deepseek and Claude) in experimentation.}

\end{enumerate}

\end{document}